\documentclass{article}

\usepackage{microtype}
\usepackage{graphicx}
\usepackage{subcaption}
\usepackage{booktabs}
\usepackage{hyperref}

\usepackage[preprint]{icml2026}

\usepackage{amsmath}
\usepackage{amssymb}
\usepackage{mathtools}
\usepackage{amsthm}
\usepackage{multirow}
\usepackage{xcolor}
\usepackage[capitalize,noabbrev]{cleveref}

\icmltitlerunning{Guardrails Beat Guidance}

\begin{document}

\twocolumn[
  \icmltitle{Guardrails Beat Guidance: \\
  A Large-Scale Study of Rules, Skills, and Persistent Configuration for Coding Agents}

  \begin{icmlauthorlist}
    \icmlauthor{Xing Zhang}{aws}
    \icmlauthor{Guanghui Wang}{aws}
    \icmlauthor{Yanwei Cui}{aws}
    \icmlauthor{Wei Qiu}{hsbc}
    \icmlauthor{Ziyuan Li}{hsbc}
    \icmlauthor{Bing Zhu}{hsbc}
    \icmlauthor{Peiyang He}{aws}
  \end{icmlauthorlist}

  \icmlaffiliation{aws}{AWS Generative AI Innovation Center}
  \icmlaffiliation{hsbc}{HSBC Holdings Plc., HSBC Technology Center, China}

  \icmlcorrespondingauthor{Peiyang He}{peiyan@amazon.com}

  \icmlkeywords{coding agents, agent rules, CLAUDE.md, agent skills, prompt engineering, guardrails, reward shaping, SWE-bench}

  \vskip 0.3in
]

\printAffiliationsAndNotice{}

\begin{abstract}
Random rules improve a coding agent's task performance as much as expert-curated ones (both $+13.8$pp on a discriminative subset of SWE-bench Verified), and in our data \emph{every} individually beneficial rule is a negative constraint (``do not refactor unrelated code''), while every individually harmful one is a positive directive (``follow code style'').
We arrive at these findings through the first large-scale controlled study of agent rule files (\texttt{CLAUDE.md}, \texttt{.cursorrules}, and the broader family of agent skills, plugin manifests, and persona definitions): we scrape 679 rule files (25{,}532 rules) from GitHub and conduct over 5{,}000 agent runs of Claude Code with Claude Opus 4.6 on SWE-bench Verified.
Three patterns emerge.
(i)~Rule \emph{polarity} cleanly separates beneficial from harmful rules; we read this through the lens of potential-based reward shaping (PBRS).
(ii)~Performance gains are largely \textbf{content-independent}: random, shuffled, mismatched-domain, and unconverted-format rule files all match curated rules, pointing to a \emph{context priming} mechanism.
(iii)~Individual rules often appear harmful in isolation yet do not visibly accumulate damage in ensemble: pass rates remain stable across rule counts from 0 to 50.
These findings expose a hidden reliability risk in the rapidly growing ecosystem of community-authored rules and skills, and they yield a clear principle for safer agent configuration: \emph{constrain what agents must not do, rather than prescribing what they should.}
\end{abstract}

\section{Introduction}
\label{sec:intro}

The rise of AI coding agents (Claude Code \citep{anthropic2025claudecode}, SWE-agent \citep{yang2024sweagent}, Devin \citep{devin2024}, OpenDevin \citep{wang2024opendevin}) has created a new mode of human--AI collaboration: rather than retraining a model, developers persistently shape its behavior through natural-language artifacts loaded at session start.
The most visible examples are rule files (\texttt{CLAUDE.md}, \texttt{.cursorrules}, \texttt{AGENTS.md}), but the same mechanism underlies a rapidly broadening family of configuration artifacts: \emph{agent skills}, plugin manifests, system-prompt overlays, and persona definitions, all of which encode what amount to ``rules'' the model is asked to follow.
These artifacts contain instructions such as ``make minimal, targeted changes,'' ``run tests before committing,'' or ``do not modify unrelated files.''
Collectively, they form a third channel of behavioral shaping, sitting between the base model and the user prompt.

This practice is now ubiquitous.
We scrape 679 rule files from GitHub (\Cref{sec:corpus}), extracting 25{,}532 individual rules across six categories.
The ecosystem is also evolving beyond manual authoring: EvolveR \citep{evolver2026} distills strategic principles from agent trajectories into reusable context, DSPy \citep{khattab2023dspy} optimizes prompt parameters, and recent analysis of iterative generative optimization \citep{igo2026} identifies rules as a critical \emph{starting artifact} that determines the reachable solution space.
Yet a fundamental question remains unanswered: \textbf{do these rules actually help, or can they silently harm?}
And if so, \emph{what properties make a rule safe versus harmful?}
Without empirical answers, both human authors and automated optimizers risk degrading the very agents they configure.

Agent rules differ from standard prompts in important ways: they are persistent (loaded every session), authored by third parties (not the model developer), and intended to shape multi-step tool-using behavior rather than single-turn generation.
As base models grow stronger through RL training \citep{composer2026}, it becomes increasingly important to know which rules still add value and which have already been absorbed into model capabilities.

We answer these questions empirically and interpret them conceptually.
\textbf{Empirically}, we run controlled experiments with a state-of-the-art coding agent on SWE-bench Verified \citep{jimenez2024swebench}, a curated set of 500 real GitHub issues.
A paired within-subject design on 58 discriminative tasks lets us measure how rule sources, counts, types, and combinations affect pass rates across over 5{,}000 agent runs.
\textbf{Conceptually}, we read the results through the lens of \emph{potential-based reward shaping} (PBRS) from reinforcement learning \citep{ng1999policy}, which provides useful vocabulary for distinguishing auxiliary signals that preserve optimal behavior (\emph{shaping}) from those that distort it.
We treat PBRS as a generative metaphor rather than a formal theory: it suggests testable hypotheses about state-dependence, accumulation, and additivity, and where its predictions fail, the failures themselves localize where the analogy ends.

\begin{figure*}[t]
\centering
\includegraphics[width=\textwidth]{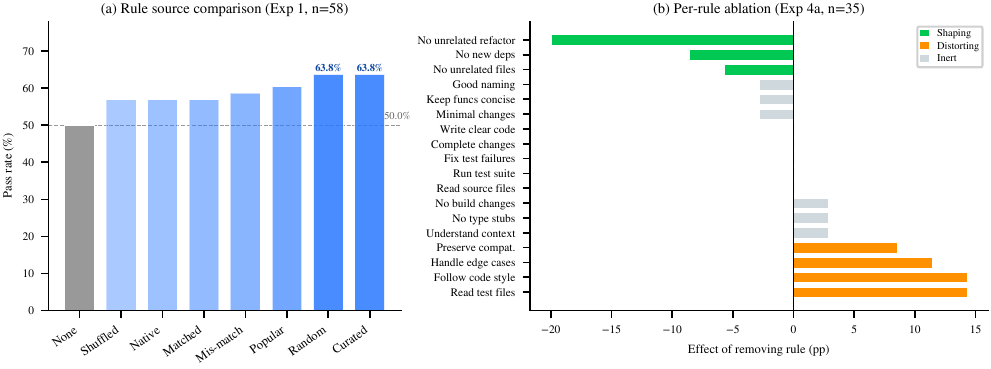}
\caption{\textbf{Headline findings.} (a)~All seven rule conditions outperform the no-rule baseline by 7--14pp, yet random rules match curated rules (both 63.8\%), pointing to a \emph{context priming} effect rather than content-specific instruction. (b)~Per-rule ablation on 18 curated rules: shaping rules (green, removal hurts) are all negative constraints; distorting rules (orange, removal helps) are all positive directives.}
\label{fig:hero}
\end{figure*}

Our headline findings are previewed in \Cref{fig:hero}.
First, rules improve performance, but their specific content matters far less than their mere presence (\Cref{fig:hero}a).
Second, per-rule ablation reveals a clean separation along rule \emph{polarity}: \emph{negative constraints} shape behavior, while \emph{positive directives} distort it (\Cref{fig:hero}b).
A third, complementary result is that performance is resilient to rule count, with no degradation up to 50 rules (\Cref{fig:scaling}a).

Our contributions are:
\begin{enumerate}
\item Evidence that rule \textbf{polarity} separates beneficial from harmful rules: in our setting, every individually beneficial rule is a negative constraint (``do not X'') and every distorting one is a positive directive (``do X'') (\Cref{sec:exp4}).
\item The first controlled evaluation of agent rule files at scale, showing that rules help primarily through \textbf{context priming} rather than specific behavioral instruction. This implies that automated rule optimizers should target \emph{structural} properties rather than semantic content (\Cref{sec:exp1}).
\item A characterization of \textbf{ensemble resilience}: individual rules that appear harmful in isolation do not accumulate damage in ensemble, with stable pass rates from 0 to 50 rules (\Cref{sec:exp2}).
\item A corpus analysis of 679 agent rule files (25{,}532 rules) from GitHub, providing the first taxonomy of this emerging configuration artifact (\Cref{sec:corpus}).
\end{enumerate}

We are explicit about scope: all empirical claims are for Claude Code with Claude Opus 4.6 on SWE-bench Verified, and the patterns we observe should be tested on other agents and benchmarks before being treated as universal.
We discuss external validity in \Cref{sec:discussion}.

\section{Related Work}
\label{sec:related}

\paragraph{Coding agents and persistent configuration.} SWE-agent \citep{yang2024sweagent}, Devin \citep{devin2024}, and OpenDevin \citep{wang2024opendevin} represent the current generation of autonomous coding agents.
All accept persistent configuration through system prompts, rule files, or, increasingly, packaged ``skills'' and plugins that bundle instructions with optional code.
The effect of these configurations on end-task performance has not been systematically studied.
\citet{arize2025rules} reported over 15\% accuracy improvement from iteratively optimized rules on SWE-bench Lite using the Cline agent; our results are consistent in direction.
We additionally disentangle the contributions of context priming versus rule content and identify which rule properties matter most.

\paragraph{Skills, plugins, and rules as a common artifact.} Recent platforms expose a growing set of mechanisms for shaping agent behavior: agent skills (markdown bundles loaded on demand), plugins or sub-agents (named capabilities with their own prompts), tool descriptions, and rule files.
Despite different surface forms, all are natural-language artifacts that prepend persistent instructions to the model's context window.
Our findings about polarity, priming, and ensemble resilience therefore apply, in principle, to any of these channels; rule files are simply the cleanest setting in which to study them at scale because the corpus is public and the injection mechanism is uniform.

\paragraph{Prompt engineering and optimization.} Extensive work has studied how prompt content affects LLM behavior \citep{reynolds2021prompt, wei2022chain, zhou2023large}.
\citet{min2022rethinking} found that in-context learning benefits more from demonstration format than content, and \citet{webson2022prompt} showed that irrelevant prompts can be as effective as relevant ones; both motivate our priming hypothesis.
Automatic optimization methods like DSPy \citep{khattab2023dspy} and APE \citep{zhou2023large} search for optimal prompts programmatically.
As we show in \Cref{sec:exp1}, the returns to \emph{content} optimization may be limited for agent rule files; what matters more is the \emph{type} of rule (constraint vs.\ directive).

\paragraph{Agent scaffold optimization.} A growing body of work optimizes agent systems through scaffolding, prompts, and rules rather than (or in addition to) modifying model weights \citep{selfevolving2025}.
EvolveR \citep{evolver2026} distills reusable strategic principles from agent interaction trajectories and injects them into agent context, a dynamic analogue of the static rule files we study.
\citet{igo2026} analyze iterative generative optimization and identify three critical design decisions, noting that the \emph{starting artifact} (initial code, prompt, or rules) determines the reachable solution space.
Our findings (\Cref{sec:experiments}) directly inform these systems: if the priming effect dominates, auto-optimizers need not search the full semantic space; if constraints outperform directives, the search narrows to negative rules; and if ensemble resilience holds, greedy single-rule optimization may miss emergent collective effects.

\paragraph{Reward shaping as conceptual lens.} \citet{ng1999policy} established that potential-based reward shaping preserves optimal policies in RL.
MiRA \citep{mira2026} uses milestone-based subgoal rewards to improve long-horizon agent RL training, and PRS \citep{prs2025} proposes progressive shaping from simple to complex objectives (e.g., formatting compliance before answer correctness).
We \emph{borrow} from the PBRS framework analogically (\Cref{sec:pbrs}) rather than claim a formal theory: it provides useful vocabulary (\emph{shaping} vs.\ \emph{distortion}, \emph{state-dependent} vs.\ \emph{state-independent}) and three falsifiable predictions about what rules should do.
Where the predictions hold, the analogy guides interpretation; where they fail, the failures themselves delineate a boundary between RL-style policy updates and LLM in-context instruction following.

\section{Rule File Corpus}
\label{sec:corpus}

To characterize the ecosystem of agent rule files, we scrape GitHub using the Code Search API for files named \texttt{CLAUDE.md}, \texttt{.cursorrules}, and related variants.

\paragraph{Collection.} We collect 679 files: 486 \texttt{CLAUDE.md} files from GitHub Code Search and 193 \texttt{.cursorrules} files from the \texttt{awesome-cursorrules} repository. Non-\texttt{CLAUDE.md} files are normalized to a common format using an LLM-based converter.

\paragraph{Extraction and cleaning.} We extract 42{,}199 individual rules, then filter to 25{,}532 after removing short fragments ($<$30 characters), duplicates, and capping at 200 rules per repository to prevent source domination.

\begin{figure}[t]
\centering
\includegraphics[width=\columnwidth]{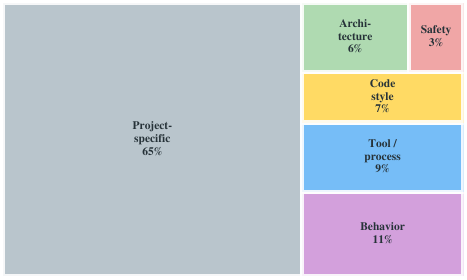}
\caption{\textbf{Taxonomy of 25{,}532 rules} from 679 GitHub rule files, classified by an LLM classifier (87\% agreement with human labels). Project-specific rules dominate (64.9\%); our experiments use the five transferable categories.}
\label{fig:taxonomy}
\end{figure}

\paragraph{Taxonomy.} Each rule is classified into one of six categories using an LLM classifier (validated at 87\% agreement with human labels on 200 rules). As shown in \Cref{fig:taxonomy}, the distribution is: \textbf{project-specific} (64.9\%), repository-level context like ``API key is in .env''; \textbf{behavior/persona} (10.8\%), cognitive directives like ``think step by step''; \textbf{tool/process} (8.9\%), workflow instructions like ``run tests before committing''; \textbf{code style} (6.5\%), formatting conventions; \textbf{architecture} (5.8\%), structural guidance; and \textbf{safety} (3.0\%), negative constraints like ``do not modify unrelated files.''

The dominance of project-specific rules (64.9\%) confirms that practitioners primarily use rule files for injecting repository context rather than behavioral guidance.
For our experiments, we focus on the five \emph{transferable} categories (excluding project-specific), which are the rules that could plausibly generalize across repositories.

\section{Conceptual Lens: PBRS}
\label{sec:pbrs}

We use potential-based reward shaping (PBRS) \citep{ng1999policy} as a \emph{conceptual lens} rather than a formal theory.
The central insight of PBRS is that auxiliary signals depending on the agent's \emph{state} preserve optimal behavior (\emph{shaping}), while state-independent signals can arbitrarily \emph{distort} it.
Translating this analogically to natural-language rules:

\begin{itemize}
\item \textbf{State-dependent rules} (e.g., ``if tests fail, fix the root cause'') condition on the agent's current situation, paralleling valid potential-based shaping. These are typically tool/process rules that specify \emph{when} to take \emph{what} action.
\item \textbf{State-independent rules} (e.g., ``write clear code'') apply a constant bias regardless of context, paralleling non-potential-based perturbation. These include style and architecture rules that apply uniformly.
\end{itemize}

This analogy generates three falsifiable predictions:
\begin{itemize}
\item[\textbf{P1}] State-dependent rules (tool/process) should help more than state-independent ones (style/architecture).
\item[\textbf{P2}] Constant-bias rules should accumulate and degrade performance at scale (a phase transition).
\item[\textbf{P3}] Shaping terms should compose additively: the effect of rules A+B should approximate the sum of their individual effects.
\end{itemize}

We test each in \Cref{sec:experiments}.
Our goal is not to prove that LLM agents implement PBRS, but to use PBRS as a structured set of expectations: where predictions hold, the vocabulary becomes useful; where they fail, the failures expose qualitative differences between RL-style policy updates and in-context instruction processing.

\section{Experimental Setup}
\label{sec:setup}

\paragraph{Scope.} All experiments use a single agent (Claude Code) with a single backbone (Claude Opus 4.6 via AWS Bedrock) on a single benchmark (SWE-bench Verified, Python bug-fix tasks).
We discuss external validity in \Cref{sec:discussion}; readers should treat findings as scoped to this configuration unless otherwise noted.

\paragraph{Agent and benchmark.} Claude Code with Claude Opus 4.6 is among the top-performing coding agent systems at the time of writing.
SWE-bench Verified \citep{jimenez2024swebench} consists of 500 curated real GitHub issues; the agent must produce a patch that passes the repository's test suite.
Evaluation uses the official SWE-bench Docker harness with per-repository containers for reproducible dependencies and Python versions.

\paragraph{Rule injection.} Claude Code automatically reads a \texttt{CLAUDE.md} file from the working directory at session start.
We swap this file before each run, varying only the rule contents while holding the agent, model, prompt template, and tooling fixed.
This requires zero modifications to the agent and mirrors the real-world authoring pattern.

\paragraph{Discriminative task selection.} Running all 500 SWE-bench Verified tasks is prohibitively expensive (${\sim}\$5$ per task-condition).
Pilot runs confirmed that 47\% of tasks are always solved and 27\% never solved regardless of rules, contributing no statistical signal.
We screen all 500 tasks with three baseline repetitions each (1{,}500 screening runs) and retain 58 \emph{discriminative} tasks (those solved 1 or 2 out of 3 times, i.e., 30--70\% baseline pass rate), where rules have room to help or hurt.

\paragraph{Paired design.} All conditions within an experiment use the same task set (58 tasks for Experiments~1--3; 35 for Experiment~4), enabling paired statistical comparisons.
We report McNemar's exact test \citep{mcnemar1947note} for pairwise comparisons and Cochran's Q test \citep{cochran1950comparison} for multi-condition comparisons.

\paragraph{Experiment summary.} Our four experiments total 3{,}544 agent runs:
Experiment~1 (rule sources: 8 conditions $\times$ 58 tasks = 464 runs),
Experiment~2 (rule counts: 19 conditions $\times$ 58 tasks = 1{,}102 runs),
Experiment~3 (rule types: 5 conditions $\times$ 58 tasks = 290 runs), and
Experiment~4 (ablation, distortion, superposition: 1{,}688 runs on 35 tasks).
Including the 1{,}500 screening runs, the total exceeds 5{,}000 agent runs at an API cost of approximately \$2{,}000.

\section{Experiments and Results}
\label{sec:experiments}

\subsection{Experiment 1: Do Rules Help?}
\label{sec:exp1}

We compare eight conditions on 58 tasks (\Cref{fig:hero}a): a no-rule \emph{baseline} and seven rule conditions chosen to probe distinct properties of rule quality:
\emph{curated} (hand-crafted best-practice rules),
\emph{random} (three randomly-drawn community rule files),
\emph{popular} (top-starred community rules),
\emph{matched} (rules from the same programming domain as the task),
\emph{mismatched} (wrong-domain rules, e.g., React rules for Python tasks),
\emph{native-only} (\texttt{CLAUDE.md} files without format conversion),
and \emph{shuffled} (the popular set with sentences randomly reordered to destroy semantic coherence).

\paragraph{Finding 1: Rules help; specific content matters far less than presence.}
Every rule condition outperforms the 50.0\% no-rule baseline by 6.9--13.8pp (\Cref{fig:hero}a, \Cref{tab:exp1_full}), yet no condition is significantly different from any other (Cochran's $Q = 4.70$, $p = 0.697$).
The closest pairwise contrast is random vs.\ baseline (McNemar $p = 0.077$; 12 tasks helped, 4 hurt).
Three patterns reinforce that the gain is largely content-independent:
\textbf{random rules tie with curated rules} at 63.8\%;
mismatched (wrong-domain) rules slightly outperform matched (same-domain) rules (58.6\% vs.\ 56.9\%);
and shuffling sentences in the popular set yields no significant change (56.9\% vs.\ 60.3\%), in contrast to known order sensitivity in few-shot in-context learning \citep{lu2022fantastically}, a contrast we read as further evidence that the rule-file mechanism is priming-based rather than demonstration-based.

No individual McNemar comparison reaches $p < 0.05$ at $n=58$; the headline rests on \emph{direction}, not magnitude.
All seven non-baseline conditions improve over baseline, and a binomial sign test on the seven directions gives $p = 0.008$.
This is modest but consistent evidence that rules genuinely help in this setting; it is not evidence that any specific rule set is reliably better than another.

The pattern is consistent with a \emph{context priming} mechanism: the mere presence of structured, on-topic instruction nudges the agent into a more careful problem-solving mode, with secondary sensitivity to content.
This parallels findings by \citet{min2022rethinking} that in-context learning benefits from demonstration \emph{format} more than \emph{content}, and \citet{webson2022prompt}'s observation that even semantically irrelevant prompts can be as effective as relevant ones.

\subsection{Experiment 2: Does Rule Count Matter?}
\label{sec:exp2}

P2 predicts that constant-bias rules accumulate and degrade performance.
We test rule counts of 0, 1, 3, 5, 10, 20, and 50 rules, each sampled with three different random seeds ($n = 58$ tasks per condition).

\paragraph{Finding 2: No phase transition; ensemble resilience.}
Pass rates stay in the 59--67\% band regardless of rule count (\Cref{fig:scaling}a), with 50 rules slightly \emph{outperforming} zero rules (66.7\% vs.\ 60.3\%).
This refutes P2 in our setting: non-potential-based biases do not visibly accumulate.
The dominant source of variance is \emph{which} rules are sampled (seed variance up to 17pp at count $=1$), not \emph{how many}.
Cochran's $Q = 22.12$ ($p = 0.227$) across all 19 conditions, confirming no significant difference.

We interpret this as \textbf{ensemble resilience}: even though individual rules may distort behavior in isolation (\Cref{sec:exp4}), their effects do not compound when stacked, suggesting the model treats the rule file as a noisy signal from which it extracts a general behavioral prior.
The practical takeaway is that, within the range we tested, developers need not aggressively prune their rule files.

\subsection{Experiment 3: Rule Type Effects}
\label{sec:exp3}

P1 predicts that state-dependent rules outperform state-independent ones.
We compare five rule types (10 rules each, matched for count) on 58 tasks (\Cref{app:exp3}).

\begin{figure*}[t]
\centering
\includegraphics[width=\textwidth]{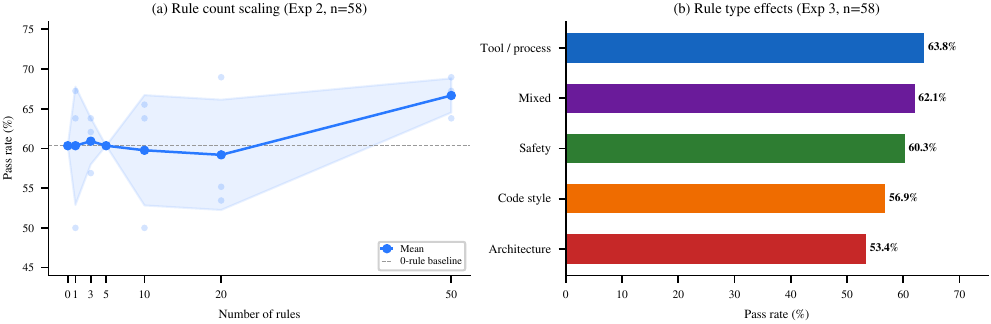}
\caption{\textbf{Rule quantity and type} ($n=58$ tasks each). (a)~Pass rate vs.\ rule count: individual seeds (dots), mean $\pm$ std (line with shading). No phase transition; 50 rules slightly outperform 0 rules. (b)~Pass rate by rule type. State-dependent tool/process rules score highest; state-independent architecture rules score lowest (10.4pp spread), consistent with P1.}
\label{fig:scaling}
\end{figure*}

\paragraph{Finding 3: Directional support for P1.}
Tool/process rules (state-dependent, analogous to valid potential-based shaping) achieve the highest pass rate (63.8\%), while architecture rules (state-independent) score lowest (53.4\%), a 10.4pp spread (\Cref{fig:scaling}b).
The ordering is consistent with P1: rules that condition on the agent's current situation (``if tests fail, investigate the root cause'') are more helpful than generic structural advice (``use the repository pattern'').
The spread does not reach statistical significance at $n=58$, so we frame this as directional rather than confirmatory evidence.

\subsection{Experiment 4: Shaping vs.\ Distorting}
\label{sec:exp4}

We perform three fine-grained analyses on 35 discriminative tasks using a curated set of 18 rules (\Cref{app:rules}), aiming to characterize \emph{which} individual rules help or hurt and \emph{how} they combine.

\paragraph{4a: Per-rule ablation.} Starting from the full 18-rule curated set (65.7\% pass rate), we remove each rule individually and measure the change (\Cref{fig:hero}b).
Using a $|\Delta| > 5$pp threshold, of 18 rules, 3 are \emph{shaping} (removal hurts), 4 are \emph{distorting} (removal helps), and 11 are \emph{inert}.

\begin{table}[t]
\caption{\textbf{Non-inert rules from per-rule ablation} ($n=35$ tasks; cf.~\Cref{fig:hero}b). \emph{Shaping}: removal hurts performance; \emph{distorting}: removal helps. $p$-values from McNemar's exact test. The 11 inert rules ($|\Delta| \leq 5$pp) are omitted; see \Cref{app:exp4a} for the complete table.}
\label{tab:exp4a}
\centering
\small
\begin{tabular}{@{}lcccc@{}}
\toprule
Rule & Polarity & Type & Effect & $p$ \\
\midrule
No unrelated refactor & ``don't'' & safety & shaping & .016 \\
No new dependencies & ``don't'' & safety & shaping & .508 \\
No unrelated files & ``don't'' & safety & shaping & .688 \\
\midrule
Read test files & ``do'' & process & distorting & .227 \\
Follow code style & ``do'' & style & distorting & .180 \\
Handle edge cases & ``do'' & style & distorting & .289 \\
Preserve compat. & ``do'' & safety & distorting & .508 \\
\bottomrule
\end{tabular}
\end{table}

\paragraph{Finding 4: Polarity separates beneficial from harmful rules.}
The pattern in \Cref{fig:hero}b and \Cref{tab:exp4a} is striking: all three shaping rules are \emph{negative constraints} (``do not X''), while all four distorting rules are \emph{positive directives} (``do X'').
A 2$\times$2 Fisher's exact test on this categorical separation (3 negatives all shaping, 4 positives all distorting) yields $p = 0.029$.
The most impactful single rule, ``do not refactor unrelated code,'' shows a 20pp drop when removed (McNemar $p = 0.016$); this is the only individually significant comparison and would not survive a strict multiple-comparison correction across all 18 rules, so we present the polarity pattern, not the per-rule $p$-values, as the primary result.

\paragraph{Threshold sensitivity.}
The 5pp shaping/distorting threshold is a choice; we therefore recompute on the same data at $\{3, 5, 7, 10\}$pp.
The partition is identical at 3pp and 5pp (3 vs.\ 4, Fisher $p = 0.029$), weakens at 7pp ($p = 0.067$), and dissolves at 10pp ($p = 0.25$); a threshold-free Mann--Whitney test on continuous $\Delta$ gives $p \approx 0.13$.
The polarity signal is thus robust to small-threshold choices but rests on a discrete partition; we treat it as suggestive evidence motivating cross-agent replication rather than a strong individual-paper claim.

We read the polarity pattern through the PBRS lens.
Positive directives such as ``read test files'' or ``follow code style'' add prescribed activities that compete for attention with the core task, analogous to reward distortion, where a misaligned auxiliary reward diverts the policy away from the true objective.
Negative constraints such as ``do not refactor unrelated code'' \emph{prune} the action space, eliminating costly deviations only when the agent is about to err, analogous to state-dependent shaping that activates exactly when needed.
The base agent already possesses strong coding capabilities; it does not need to be told \emph{what} to do, but it benefits from being told what \emph{not} to do.

\paragraph{4b: Distortion on previously-solved tasks.}
We isolate 17 tasks that the baseline agent solves reliably (88.2\% on re-run) and apply each of the 18 rules \emph{individually} to measure how often a rule breaks a task the agent already solved.

\begin{figure*}[t]
\centering
\includegraphics[width=\textwidth]{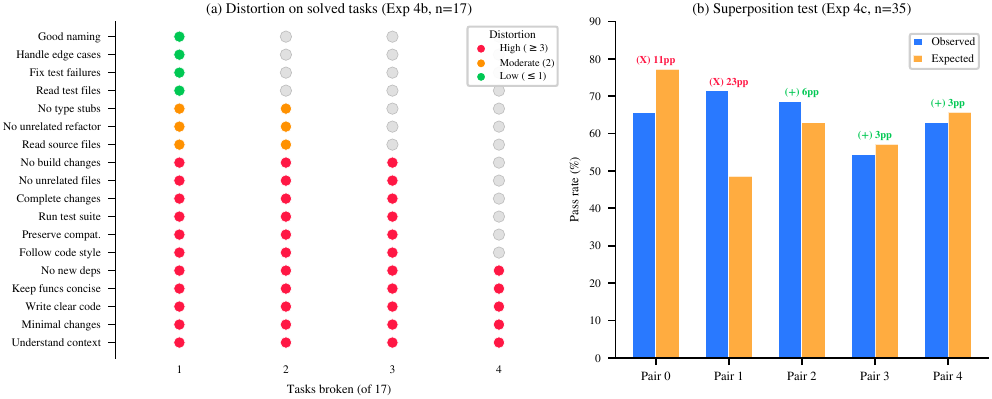}
\caption{\textbf{Individual rule effects} ($n=35$ tasks). (a)~Per-rule distortion: each of the 18 curated rules applied individually to 17 baseline-solved tasks. 14/18 rules break $\geq$2 tasks; the worst break 4/17. (b)~Superposition test on 5 rule pairs: observed vs.\ additive-expected pass rate. Three pairs ($+$) are approximately additive; two ($\times$) show strong non-linearity (max error 22.9pp).}
\label{fig:detail}
\end{figure*}

As shown in \Cref{fig:detail}a, 14 of 18 rules break at least 2 previously-solved tasks; the worst offenders (``understand full context,'' ``keep functions concise,'' ``do not install dependencies'') each break 4/17 (24\%).
Only 4 rules are safe ($\leq 1$ task broken).
We treat this evidence as suggestive rather than conclusive: with a baseline retest reliability of 88.2\%, each task has an $\approx 11.8\%$ chance of flipping under run-to-run variance alone, so any single rule's break count cannot be read as fully causal.

\paragraph{Finding 5: Individual harm does not compound in ensemble.}
The robust comparison is between any single rule and the ensemble.
If individual rules were genuinely as distortive as 4b suggests, stacking 18 of them should compound the damage; instead, the 18-rule ensemble holds at 65.7\% (comparable to the per-pair no-rule baselines of 68.6--74.3\% measured on the same 35 tasks in 4c).
We interpret the gap between individual-rule appearance and ensemble behavior as \emph{distortion averaging}: idiosyncratic biases partially cancel when stacked, while the shared priming signal persists.
This implies that greedy single-rule evaluation is the wrong lens for selecting rules: rules that look harmful in isolation may be useful, or at least neutral, in combination.

\paragraph{4c: Superposition test (P3).}
For five randomly-selected rule pairs from different categories, we measure pass rates under four conditions: no rules, rule A only, rule B only, and A+B together.
Under PBRS-style additivity, $P(\text{A+B}) \approx P(\text{A}) + P(\text{B}) - P(\text{none})$.

\paragraph{Finding 6: Partial additivity.}
Three of five pairs are approximately additive (error $<10$pp), but two show strong non-linearity (\Cref{fig:detail}b, \Cref{tab:exp4c_full}).
In the most extreme case (Pair~1), rule A alone \emph{hurts} ($-17$pp) but A+B together returns to the no-rule baseline; B fully compensates for A's distortion.
In Pair~0, rules A and B each individually help ($+3$pp, $+6$pp) but together \emph{hurt} ($-3$pp), suggesting interference.

P3 thus holds only approximately.
Rules can interact through \emph{compensation} (one rule rescues another's damage) or \emph{interference} (two helpful rules conflict), mechanisms with no direct analogue in the linear PBRS framework. This is one of the points where our empirical findings exceed the conceptual lens.

\section{Discussion}
\label{sec:discussion}

\begin{table}[t]
\centering
\small
\setlength{\tabcolsep}{4pt}
\begin{tabular}{@{}p{0.09\columnwidth}p{0.83\columnwidth}@{}}
\toprule
\multicolumn{2}{@{}l@{}}{\textbf{Box 1: Rule authoring guidelines for coding agents.}} \\
\multicolumn{2}{@{}l@{}}{\footnotesize Scoped to Claude Code-class agents on SWE-bench-style tasks.} \\
\midrule
\textbf{Do} & Write \emph{negative constraints} (``do not refactor unrelated code''). \\
\textbf{Don't} & Write positive directives (``follow code style,'' ``handle edge cases'') without auditing them. \\
\midrule
\textbf{Do} & Include rules; even random or mismatched-domain rules help. \\
\textbf{Don't} & Spend effort optimizing wording; presence beats phrasing. \\
\midrule
\textbf{Do} & Use up to 50 rules; no degradation observed. \\
\textbf{Don't} & Rely on single-rule evaluation; ensemble effects dominate. \\
\bottomrule
\end{tabular}
\end{table}

\paragraph{The priming hypothesis.}
Our most robust pattern is that rule \emph{content} matters far less than rule \emph{presence}.
Random rules match curated rules; shuffled rules match ordered; wrong-domain rules match right-domain.
We hypothesize that rule files function primarily as \textbf{context primes}: structured on-topic text that nudges the model into a more careful problem-solving mode. This parallels prompt-sensitivity findings that LLMs respond to format and framing more than semantic content \citep{webson2022prompt, min2022rethinking}.
The fact that four independent conditions (random, shuffled, mismatched, native-only) all yield similar gains makes content-specific explanations difficult to maintain.
A sharper test would compare full rule files against minimal primes (e.g., a single sentence like ``follow best practices''); we leave this to future work.

This implication is not specific to rule files.
Any system that prepends persistent instructions to the model context (\textbf{agent skills, plugins, system prompts, tool descriptions, persona definitions}) likely benefits from the same priming effect, and is exposed to the same distortion risk.
We expect the polarity asymmetry we observe in rule files to surface, in similar form, wherever a capable agent receives prescriptive natural-language configuration.
A converging signal comes from persona prompting: \citet{zhang2026alignmentfloor} report that resistance-oriented personas (e.g., a ``skeptic'' persona) reduce sycophancy on a weak model from 50\% to 5\%, while engagement-oriented personas (Extraversion, Openness) substantially increase harmful compliance.
The shape of that finding mirrors ours: prescribing what \emph{not} to lean toward stabilizes behavior, while prescribing what to lean \emph{toward} destabilizes it.

\paragraph{Guardrails beat guidance.}
The polarity pattern in Experiment~4a yields a simple deployment heuristic: \textbf{constrain what the agent must not do, rather than prescribing what it should}.
A capable base agent does not need to be told \emph{how} to code; it benefits from being told what \emph{not} to do.
This sits well with the PBRS lens, since negative constraints tend to be state-dependent (they activate only when the agent is about to err) and therefore closer to valid potential-based shaping.
A human-factors analogue is the aviation checklist, which prevents specific failure modes for pilots who already know how to fly, in contrast to a procedural manual that prescribes the entire flight.
The deployment risk is that a team adding well-intentioned directives such as ``follow code style'' or ``handle edge cases'' may inadvertently introduce regressions that are invisible to standard testing.
Auditing rule files, skills, plugins, and personas for positive directives (and prioritizing negative constraints) is therefore a low-cost intervention.
We are careful about the strength of this advice: at $n=35$, our individual-rule comparisons are statistically modest, and the recommendation rests on the polarity pattern rather than per-rule significance.

\paragraph{PBRS as a useful lens.}
The lens earns its keep on P1 (directional support) and on the polarity reading of Experiment~4a (\Cref{fig:hero}b), but it loses on P2 and P3.
The absence of a phase transition (P2) suggests that LLM behavior under stacked rules does not compound in the way RL policy updates would under stacked biased rewards, a qualitatively different mechanism.
The compensation and interference effects in 4c (P3) have no analogue in linear potential-based shaping.
We therefore treat PBRS as a generative metaphor rather than a theory: it produced three falsifiable predictions and a vocabulary that fits our cleanest empirical pattern, while its failures usefully localize where the analogy ends.

\paragraph{Implications for automated rule and skill optimization.}
Our findings inform systems that automatically optimize agent rules and skills \citep{khattab2023dspy, zhou2023large, evolver2026}.
First, the priming effect implies diminishing returns to \emph{content-level} optimization (rephrasing for clarity or specificity); optimizers should target \emph{structural} properties such as polarity, category composition, and count.
Second, the polarity pattern provides a concrete search prior: auto-generated principles should be biased toward negative constraints.
Third, ensemble resilience warns against greedy per-rule selection: rules judged harmful in isolation often have their effects absorbed in ensemble, and the ``starting artifact'' \citep{igo2026} that an optimizer inherits may dominate any local edits.
The flip side of ensemble resilience is a lifecycle problem: as self-evolving skill libraries accumulate skills over time, retrieval and routing can degrade even when each individual skill is harmless, a failure mode characterized by \citet{zhang2026librarydrift} as \emph{library drift}.
Constraint-based polarity priors and library-drift governance are complementary: the former biases what gets added, the latter governs what gets retired.

\paragraph{The shelf life of rules.}
As base models grow stronger through RL training \citep{composer2026}, rules describing \emph{how} to code (style, formatting, generic process) are increasingly internalized into model capabilities, while rules encoding \emph{what not to do} (project-specific scope constraints, safety prohibitions) cannot be absorbed by training because they reflect human or organizational priorities the model has no way to know.
Our data is consistent with this asymmetry: all three shaping rules in \Cref{tab:exp4a} are negative constraints (``do not refactor unrelated code,'' ``do not install new dependencies,'' ``do not modify unrelated files''), and the distorting rules consist mainly of style and process directives the agent likely already has internalized.
We tentatively expect the long-term value of rule files (and skills) to lie not in teaching agents \emph{how} to do their work, but in encoding the safety constraints and organizational values that pretraining and RL cannot easily absorb.

\paragraph{Practical recommendations.}
For practitioners configuring Claude Code-class agents, we summarize the takeaways in Box~1 and elaborate them here:
\begin{enumerate}
\item \textbf{Prefer ``do not'' over ``do.''} Negative constraints are the most consistently beneficial rule type in our data and the most likely to remain valuable as base models improve.
\item \textbf{Audit positive directives.} ``Follow code style,'' ``handle edge cases,'' and similar well-intentioned directives can distract a capable agent from the core task.
\item \textbf{Presence beats phrasing.} Any reasonable set of on-topic instructions activates the priming effect; do not over-engineer wording.
\item \textbf{Do not over-prune; do not over-trust single rules.} Up to 50 rules shows no degradation, and rules that look harmful in isolation often have their distortion absorbed in ensemble.
\end{enumerate}

\paragraph{Limitations.}
\emph{Single-vendor, single-model, single-benchmark.}
All experiments use Claude Code with Claude Opus 4.6 on SWE-bench Verified (Python bug fixes only).
Weaker models may rely more heavily on rule guidance and could show different polarity sensitivities; benchmarks that reward broader changes (rather than minimal patches) may not favor ``do not refactor''-style constraints in the same way.
The one external data point, \citet{arize2025rules}'s $>$15\% improvement from rules on SWE-bench Lite with the Cline agent, is consistent with our priming claim but does not generalize the polarity claim.
Cross-agent and cross-task replication is the most important next step.

\emph{Statistical power and baseline variance.}
At $n=58$ (Experiments~1--3) and $n=35$ (Experiment~4), our McNemar tests detect $\sim$15pp paired effects; the 14pp average rule gain in Experiment~1 does not cross $p < 0.05$ at the per-condition level, and we rely on a binomial sign test across seven conditions ($p = 0.008$) for the headline.
The no-rule baseline also differs between Experiment~1 (50.0\%, single trajectory per task) and Experiment~2 (60.3\%, averaged over three seeds); part of this gap reflects the run-to-run stochasticity of single-trajectory agent evaluation, which we mitigate in Experiment~2 by seed-averaging.
For Experiment~4a, the only individually significant rule (rule 6, $p=0.016$) would not survive a strict multiple-comparison correction across 18 tests; we accordingly present polarity, not per-rule significance, as the primary finding.
For Experiment~4b, the 88.2\% baseline retest reliability bounds how strongly individual ``breaks'' can be interpreted; the more robust contrast is the gap between any single rule and the ensemble.

\emph{Selection bias.}
By selecting discriminative tasks (30--70\% baseline pass rate), we enrich for tasks where rules can produce a measurable effect.
Screening data confirms that rules do not degrade the remaining tasks (always-pass and always-fail tasks behave identically across conditions), so the selection bias does not mask hidden harm in the easy/hard tails, though it does limit our ability to characterize the full distribution.

\emph{Future work.}
Priority extensions are cross-agent and cross-task replication, sensitivity analyses for the polarity claim under richer multiple-comparison corrections, constraint-aware rule and skill evolution \citep{evolver2026}, and converting text rules into auxiliary RL reward signals \citep{prs2025, mira2026}.

\section{Conclusion}
\label{sec:conclusion}

Agent rule files are the first widely-adopted mechanism through which developers persistently shape AI agent behavior, and they are now joined by skills, plugins, and other natural-language configuration artifacts.
Our study of 25{,}532 rules across 679 files, evaluated through over 5{,}000 runs of Claude Code with Claude Opus 4.6 on SWE-bench Verified, surfaces three patterns: rule polarity appears to separate beneficial negative constraints from distorting positive directives; gains are largely content-independent, pointing to context priming; and rule ensembles remain resilient even at 50 rules.
We frame these as scoped empirical findings rather than universal laws, but if they hold up under cross-agent replication, they offer a concrete principle for safer agent configuration: one that applies to any channel through which humans hand persistent instructions to a capable model.
As base models grow stronger, the enduring value of such artifacts is likely to lie not in teaching agents \emph{how} to do their work, but in encoding the safety constraints and organizational values that training alone cannot easily capture.

\section*{Impact Statement}

This paper studies how natural-language rule files affect AI coding agent behavior and reliability.
Our finding that well-intentioned rules can silently degrade agent performance has direct safety implications for the growing ecosystem of community-authored rules, skills, and plugins; the actionable guidance we provide (prefer negative constraints over positive directives) helps practitioners configure agents more safely.
All experiments are conducted in a controlled benchmarking setting and we see no specific ethical concerns beyond those generally associated with improving AI coding agents.

\clearpage
\appendix

\section{Curated Rule Set}
\label{app:rules}

The 18 rules used in Experiments 4a--4c, hand-crafted to represent best practices across five transferable categories (excluding project-specific). Rules 1--3: understanding/context (process); 4: scope (process); 5: style; 6, 15--18: negative constraints (safety); 7: compatibility (safety); 8--9: testing (process); 10--11: code quality (style); 12: edge cases (style); 13: structure (architecture); 14: naming (style).

\begin{enumerate}\itemsep0pt\parskip0pt\parsep0pt
\small
\item Read the relevant source files to understand existing code structure and style.
\item Identify and read existing test files related to the code you will change.
\item Understand the full context of the issue before proposing a fix.
\item Make minimal, targeted changes that directly address the issue.
\item Follow the existing code style and conventions of the repository.
\item Do not refactor unrelated code or add unnecessary features.
\item Preserve backward compatibility unless the issue specifically requires breaking changes.
\item Run the existing test suite to verify your changes don't break anything.
\item If tests fail, investigate and fix the root cause rather than modifying tests.
\item Ensure your changes are complete; do not leave TODOs or partial implementations.
\item Write clear, readable code. Prefer explicit over clever.
\item Handle edge cases that are relevant to the issue.
\item Keep functions focused and concise.
\item Use meaningful variable and function names consistent with the codebase.
\item Do not modify files unrelated to the issue.
\item Do not add type stubs, documentation, or formatting changes unless requested.
\item Do not install new dependencies unless absolutely necessary.
\item Do not change the project's build or test configuration.
\end{enumerate}

\section{Complete Experiment 1 Results}
\label{app:exp1}

\Cref{tab:exp1_full} reports all 8 conditions from Experiment~1 with pass rates and pairwise McNemar tests vs.\ the no-rule baseline.

\begin{table}[ht]
\caption{\textbf{Experiment 1: complete results} ($n=58$ tasks). Pass rates and pairwise McNemar tests vs.\ the no-rule baseline, sorted by pass rate.}
\label{tab:exp1_full}
\centering
\small
\begin{tabular}{@{}lccccc@{}}
\toprule
Condition & Pass \% & Helped & Hurt & Disc. & $p$ \\
\midrule
Baseline (none) & 50.0 & --- & --- & --- & --- \\
\midrule
Random & 63.8 & 12 & 4 & 16 & .077 \\
Curated & 63.8 & 14 & 6 & 20 & .115 \\
Popular & 60.3 & 18 & 12 & 30 & .361 \\
Mismatched & 58.6 & 15 & 10 & 25 & .424 \\
Matched & 56.9 & 11 & 7 & 18 & .481 \\
Shuffled & 56.9 & 13 & 9 & 22 & .523 \\
Native only & 56.9 & 15 & 11 & 26 & .556 \\
\bottomrule
\end{tabular}
\vskip 0.05in
{\footnotesize Helped = tasks flipped from fail to pass; Hurt = pass to fail; Disc.\ = total discordant pairs. All $p$-values from McNemar's exact test. Cochran's $Q = 4.70$, $p = 0.697$ across all 8 conditions.}
\end{table}

\section{Complete Experiment 3 Results}
\label{app:exp3}

\begin{table}[ht]
\caption{\textbf{Experiment 3: pass rate by rule type} ($n=58$ tasks, 10 rules each). PBRS categories follow \Cref{sec:pbrs}.}
\label{tab:exp3_full}
\centering
\small
\begin{tabular}{@{}lcc@{}}
\toprule
Rule type & PBRS category & Pass \% \\
\midrule
Tool / process & State-dependent & 63.8 \\
Mixed (all types) & Mixed & 62.1 \\
Safety & Constraint-based & 60.3 \\
Code style & State-independent & 56.9 \\
Architecture & State-independent & 53.4 \\
\bottomrule
\end{tabular}
\vskip 0.05in
{\footnotesize Cochran's $Q = 3.05$, $p = 0.549$. No pairwise comparison reaches significance (all McNemar $p > 0.21$).}
\end{table}

\section{Complete Experiment 4a Results}
\label{app:exp4a}

\Cref{tab:exp4a_full} reports all 18 rules from the per-rule ablation, sorted by effect size $\Delta$. We define $\Delta$ as the change in pass rate when the rule is \emph{removed} from the 18-rule set (65.7\% baseline), sign-inverted so that positive $\Delta$ means the rule is helpful.

\begin{table}[ht]
\caption{\textbf{Experiment 4a: complete per-rule ablation} ($n=35$ tasks). Positive $\Delta$ means the rule is helpful (removal hurts); sorted by $\Delta$ descending. Threshold for \emph{shaping}/\emph{distorting}: $|\Delta| > 5$pp.}
\label{tab:exp4a_full}
\centering
\small
\begin{tabular}{@{}rlrcc@{}}
\toprule
\# & Rule (abbreviated) & $\Delta$ (pp) & Status & $p$ \\
\midrule
6 & No unrelated refactor & $+$20.0 & Shaping & .016 \\
17 & No new dependencies & $+$8.6 & Shaping & .508 \\
15 & No unrelated files & $+$5.7 & Shaping & .688 \\
4 & Minimal changes & $+$2.9 & Inert & 1.0 \\
13 & Keep funcs concise & $+$2.9 & Inert & 1.0 \\
14 & Good naming & $+$2.9 & Inert & 1.0 \\
1 & Read source files & 0.0 & Inert & 1.0 \\
8 & Run test suite & 0.0 & Inert & 1.0 \\
9 & Fix test failures & 0.0 & Inert & 1.0 \\
10 & Complete changes & 0.0 & Inert & 1.0 \\
11 & Write clear code & 0.0 & Inert & 1.0 \\
3 & Understand context & $-$2.9 & Inert & 1.0 \\
16 & No type stubs & $-$2.9 & Inert & 1.0 \\
18 & No build changes & $-$2.9 & Inert & 1.0 \\
7 & Preserve compat. & $-$8.6 & Distorting & .508 \\
12 & Handle edge cases & $-$11.4 & Distorting & .289 \\
2 & Read test files & $-$14.3 & Distorting & .227 \\
5 & Follow code style & $-$14.3 & Distorting & .180 \\
\bottomrule
\end{tabular}
\end{table}

\section{Complete Experiment 4c Results}
\label{app:exp4c}

\Cref{tab:exp4c_full} reports all 5 rule pairs from the superposition test.

\begin{table}[ht]
\caption{\textbf{Experiment 4c: superposition test} ($n=35$ tasks). Expected rate under additivity: $P_{\text{exp}} = P_A + P_B - P_{\text{none}}$. A pair is judged approximately additive if $|P_{\text{A+B}} - P_{\text{exp}}| < 10$pp.}
\label{tab:exp4c_full}
\centering
\small
\resizebox{\columnwidth}{!}{%
\begin{tabular}{@{}lccccccl@{}}
\toprule
& None & A & B & A+B & Exp. & Err. & Additive? \\
& (\%) & (\%) & (\%) & (\%) & (\%) & (pp) & \\
\midrule
Pair 0 & 68.6 & 71.4 & 74.3 & 65.7 & 77.1 & 11.4 & No \\
Pair 1 & 71.4 & 54.3 & 65.7 & 71.4 & 48.6 & 22.9 & No \\
Pair 2 & 71.4 & 65.7 & 68.6 & 68.6 & 62.9 & 5.7 & Yes \\
Pair 3 & 74.3 & 62.9 & 68.6 & 54.3 & 57.1 & 2.9 & Yes \\
Pair 4 & 74.3 & 80.0 & 60.0 & 62.9 & 65.7 & 2.9 & Yes \\
\bottomrule
\end{tabular}%
}
\end{table}

\end{document}